\documentclass{article}
\usepackage{arxiv}
\usepackage[utf8]{inputenc}
\usepackage[T1]{fontenc}

\usepackage{graphicx}
\usepackage{amsmath}
\usepackage{amssymb}
\usepackage{marvosym}
\usepackage[bottom]{footmisc}
\renewcommand{\footnoterule}{\kern-3pt\hrule width 0.28\textwidth\kern2.6pt}
\newcommand{\runinheading}[1]{{\fontsize{11}{13}\selectfont\bfseries #1}}
\usepackage{booktabs}
\usepackage[table]{xcolor}
\usepackage{multirow}
\usepackage{xspace}
\usepackage{microtype}
\usepackage{float}
\usepackage{comment}
\usepackage{flafter}
\usepackage{wrapfig}

\usepackage[font=small,labelfont=bf,justification=justified,singlelinecheck=false]{caption}
\AtBeginDocument{\raggedbottom}

\usepackage[nocompress]{cite}   % 数字引用（配合 BibTeX）
\usepackage[hidelinks]{hyperref}

\newcommand{\ours}{GeoAlign\xspace}
\newcommand{\oursfusion}{DGCA\xspace}
\newcommand{\oursbase}{GeoAlign w/o DGCA\xspace}
\newcommand{\oursfull}{GeoAlign\xspace}
\newcommand{\teacher}{Teacher\xspace}
\newcommand{\dataset}{RoboTwin\xspace}

\newcommand{\tablebodyfont}{\footnotesize}
\definecolor{BestCell}{HTML}{D9EAD3}
\definecolor{SecondCell}{HTML}{FCE5CD}
\definecolor{GainText}{HTML}{1F4E79}
\newcommand{\best}[1]{\cellcolor{BestCell}\textbf{#1}}
\newcommand{\second}[1]{\cellcolor{SecondCell}\textbf{#1}}
\newcommand{\gain}[2]{#1\hspace{0.35em}{\normalfont\color{GainText}(#2)}}
\newcommand{\tablecaptiontitle}[1]{\textbf{#1.}}
\newcommand{\figurecaptiontitle}[1]{\textbf{#1.}}
\title{
\ours: Distilling Geometric Priors for Real-Time Robot Manipulation
}

\newcommand{\paperauthors}{%
\normalfont\begin{minipage}{\textwidth}
\centering
Xiangyang Liu\textsuperscript{1*}, Haosong Peng\textsuperscript{2*},
Sicheng Tian\textsuperscript{1}, Shuhan Zhao\textsuperscript{1},\\[2pt]
Wenchao Xu\textsuperscript{2},
Yufeng Zhan\textsuperscript{1\Letter},
Yuanqing Xia\textsuperscript{1}\\[6pt]
{\small\textsuperscript{1}Beijing Institute of Technology\\
\textsuperscript{2}The Hong Kong University of Science and Technology}

\end{minipage}%
}

\makeatletter
\renewcommand{\@maketitle}{%
  \par\begingroup
  \centering
  \hrule height 2pt\relax
  \vskip 12pt
  {\LARGE\scshape\@title\par}
  \vskip 12pt
  \hrule height 2pt\relax
  \vskip 14pt

  \vskip 12pt
  \endgroup
}
\makeatother
\begin{document}

\maketitle
\noindent\paperauthors\par
\vspace{5pt}
{\centering\normalsize Email: \href{mailto:yu-feng.zhan@bit.edu.cn}{\textcolor{blue}{\texttt{yu-feng.zhan@bit.edu.cn}}}\par}
\medskip
\begingroup
\renewcommand{\thefootnote}{}
\footnotetext{\textsuperscript{*}Equal Contribution. \textsuperscript{\Letter}Corresponding Author.}
\endgroup

\begin{figure}[H]
\centering
\includegraphics[width=1\linewidth]{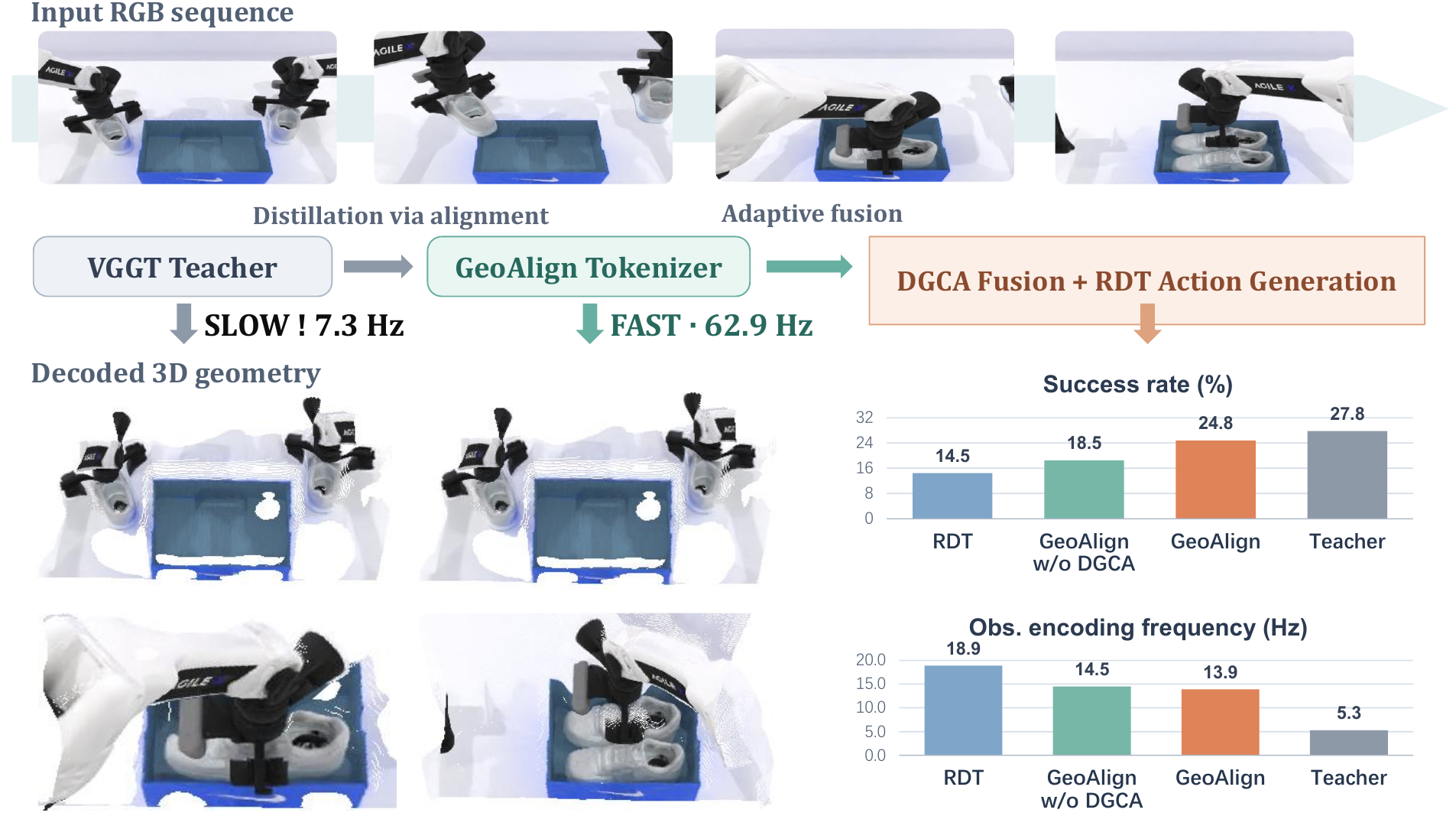}
\caption{\figurecaptiontitle{Real-time 3D geometry tokenizer} GeoAlign distills and
aligns VGGT features, while DGCA fuses them with RDT for action generation.
Results show geometric fidelity, task performance, and efficiency.}
\label{fig:teaser}
\end{figure}

% ---------------------------------------------------------------------------
% Sections
% ---------------------------------------------------------------------------
\begin{abstract}
Imitation-learning policies for robotic manipulation largely rely on 2D visual representations with limited spatial awareness.
Large 3D vision foundation models can recover rich geometric priors directly from multi-view RGB observations, but their high computational cost precludes per-frame use in real-time control.
We present \ours{}, a spatial compression framework that distills the last-block features of VGGT into a lightweight, patch-aligned geometry tokenizer without additional 3D sensors.
To integrate the resulting geometry tokens with 2D semantic features, we introduce Dual-Gated Cross-Attention, which performs input-adaptive, bidirectional alignment: geometric cues highlight spatially relevant semantic features, while semantic context selects task-relevant geometry before patch-level cross-attention.
Experiments on four \dataset benchmark tasks and two real-world bimanual manipulation tasks show that \ours outperforms simple feature concatenation, retains 89.2\% of the non-deployable teacher's simulation success rate while requiring only 37.9\% of its observation-encoding time, and achieves a 75.0\% average success rate on the real robot.
\end{abstract}

% ---------------------------------------------------------------------------
% Teaser in the single-column preprint layout.
% ---------------------------------------------------------------------------

\section{INTRODUCTION}

Humans perform complex manipulation tasks through natural perception and reasoning about the 3D world, leveraging strong spatial awareness.
Current imitation-learning methods~\cite{chi2025diffusion, liu2025rdt, black2024pi_0} for manipulation, however, do not explicitly encode 3D geometry and depend heavily on 2D visual features with limited spatial awareness.
A direct remedy for insufficient perception is to augment RGB observations with additional modalities, such as depth maps or point clouds~\cite{li2026pointvla}.
Prior work has shown that injecting 3D information indeed improves manipulation performance~\cite{ze20243d}, yet such methods rely on extra sensors or calibration setups, increasing overall system complexity.
Meanwhile, these sensors are susceptible to noise on transparent, reflective, or textureless surfaces~\cite{liu2025manipulation}, limiting large-scale real-world deployment.

Recent progress in large 3D vision foundation models offers a promising avenue for obtaining 3D priors directly from RGB input~\cite{wang2025vggt, peng2026omnivggt}.
Specifically, these 3D foundation models jointly process multiple RGB views to directly predict camera parameters and dense 3D geometry, such as depth and point maps, in a single forward pass.
These models provide a promising route to equipping downstream manipulation policies with 3D spatial understanding, without requiring additional depth or point-cloud sensors.
However, their high computational cost makes per-frame inference too slow for real-time robot control, especially in multi-view settings.
For example, as illustrated in Fig.~\ref{fig:teaser}, the 1B-parameter VGGT~\cite{wang2025vggt} model takes 137 ms per frame for inference on an RTX PRO 6000, corresponding to only 7.3 Hz.
The key challenge, therefore, is to harness the rich spatial priors of large 3D vision foundation models without incurring their prohibitive inference cost, thereby enabling real-time 3D-aware manipulation.

To address this computational bottleneck, we propose \textbf{\ours}, a spatial compression framework that replaces full per-frame inference through a large 3D foundation model with a lightweight geometry tokenizer.
The tokenizer is trained to reproduce the teacher's last-block features and operates at every policy observation/replanning step, enabling efficient 3D perception for real-time manipulation.
To effectively integrate the distilled geometric features with 2D semantic features, we further propose \textbf{Dual-Gated Cross-Attention (\oursfusion)}, which enables input-adaptive, bidirectional alignment between the two feature streams.
Specifically, a geometry-conditioned gate uses geometric cues to highlight spatially relevant semantic features, while a semantic-conditioned gate uses semantic context to select task-relevant geometric features.
Cross-attention then integrates the mutually calibrated features at the patch level, allowing geometric structure and semantic context to complement each other with only modest overhead over simple concatenation.
Extensive experiments on \dataset benchmark tasks and two real-world bimanual manipulation tasks demonstrate that our approach outperforms simple feature concatenation, retains 89.2\% of the non-deployable teacher's simulation success rate using only 37.9\% of its observation-encoding time, and achieves a 75.0\% average success rate on the real robot.
In summary, our contributions are as follows:
\begin{itemize}
    \item We propose \textbf{\ours}, a spatial compression framework that distills VGGT features into compact geometry tokens for efficient 3D-aware manipulation without additional sensors.

    \item To effectively integrate visual cues, we introduce \textbf{\oursfusion}, an adaptive fusion module that enables bidirectional alignment between semantic and geometric features.

    \item Extensive experiments across simulation benchmarks and real-world bimanual manipulation tasks show that \ours achieves substantially faster feature extraction than its teacher while maintaining strong and consistent task success rates.
\end{itemize}

\section{RELATED WORK}

\subsection{2D-Based Imitation Learning Robot Manipulation Policies}
Recent advances in imitation learning have demonstrated strong policy performance using purely 2D visual input~\cite{li2024vision, black2024pi_0, team2024octo, ahn2022can, wu2024unleashing}.
Prominent methods leverage powerful visual encoders~\cite{zhai2023sigmoid, dosovitskiy2020image} and generative policies~\cite{chi2025diffusion, zhao2023learning, team2024octo, intelligence2025pi_, liu2025rdt} to learn from demonstrations.
Open-source and efficient VLA variants further scale this paradigm to broader robot data and deployment settings~\cite{kim2024openvla, wen2025dexvla, wen2025tinyvla}.
While effective on many manipulation tasks, these methods typically rely on RGB observations without explicit reasoning about the underlying 3D scene.
They therefore often struggle with tasks that require spatial reasoning, depth understanding, handling of occlusions, or fine-grained manipulation.

\subsection{3D-Aware Imitation Learning Robot Manipulation Policies}
To overcome the limitations of 2D-only perception, a variety of methods use RGB-D, point-cloud, or voxel-based representations to enhance spatial awareness in robotic tasks~\cite{ze20243d, li2026pointvla, gervet2023act3d, zhen20243d}.
Representative designs couple 3D scene representations with diffusion or attention-based policies~\cite{ke20243d, shridhar2023perceiver}, while Gaussian scene representations provide another route to geometry-aware manipulation~\cite{lu2024manigaussian}.
Such approaches, however, require additional depth sensors, which are prone to noise, particularly under challenging conditions involving transparent or reflective surfaces.
Moreover, real-world deployment may be constrained by hardware cost or limited sensor coverage.
In contrast, our work avoids explicit depth sensing and instead builds on 3D vision foundation models, extracting geometry-aware features directly from RGB input.

\subsection{Policies with Priors from Large 3D Foundation Models}
\label{sec:related3d}
Recently, pretrained 3D foundation models (e.g., DUSt3R~\cite{wang2024dust3r}, VGGT~\cite{wang2025vggt}, and SpatialVLA~\cite{qu2025spatialvla}) provide rich geometric priors~\cite{leroy2024grounding, weinzaepfel2022croco, yang2024depthv2, yin2023metric3d} for visual manipulation, but their high computational cost limits direct deployment.
Existing work addresses this issue along two directions: model compression and inference amortization.
The former compresses large 3D models into lightweight visual encoders~\cite{vuong2025improving, hinton2015distilling} to replace conventional 2D visual backbones, e.g., eVGGT; the latter reuses features in the temporal dimension, e.g., R3DP~\cite{zhang2026r3dp}, amortizing the cost by querying the slow VGGT less frequently.
Both directions substantially reduce the cost of accessing 3D priors, making geometry-aware representations more practical for real-time control.
However, existing methods largely emphasize efficient feature extraction or reuse, while the token-level alignment and integration of 3D priors with the semantic representations used by downstream manipulation policies remain less explored.
Our work complements these approaches: rather than directly using a lightweight 3D encoder as a replacement for the 2D visual backbone, \ours distills VGGT into a geometry tokenizer whose outputs are position-wise aligned with the SigLIP patch grid, enabling low-cost fusion of semantic and geometric features.
%
%Building on this aligned representation, \oursfusion uses adaptive gating and cross-attention to enable bidirectional interaction between semantic and geometric features, allowing the policy to selectively exploit task-relevant spatial information.

\section{METHOD}
\label{sec:method}

\begin{figure}[!htbp]
\centering
\includegraphics[width=1\textwidth]{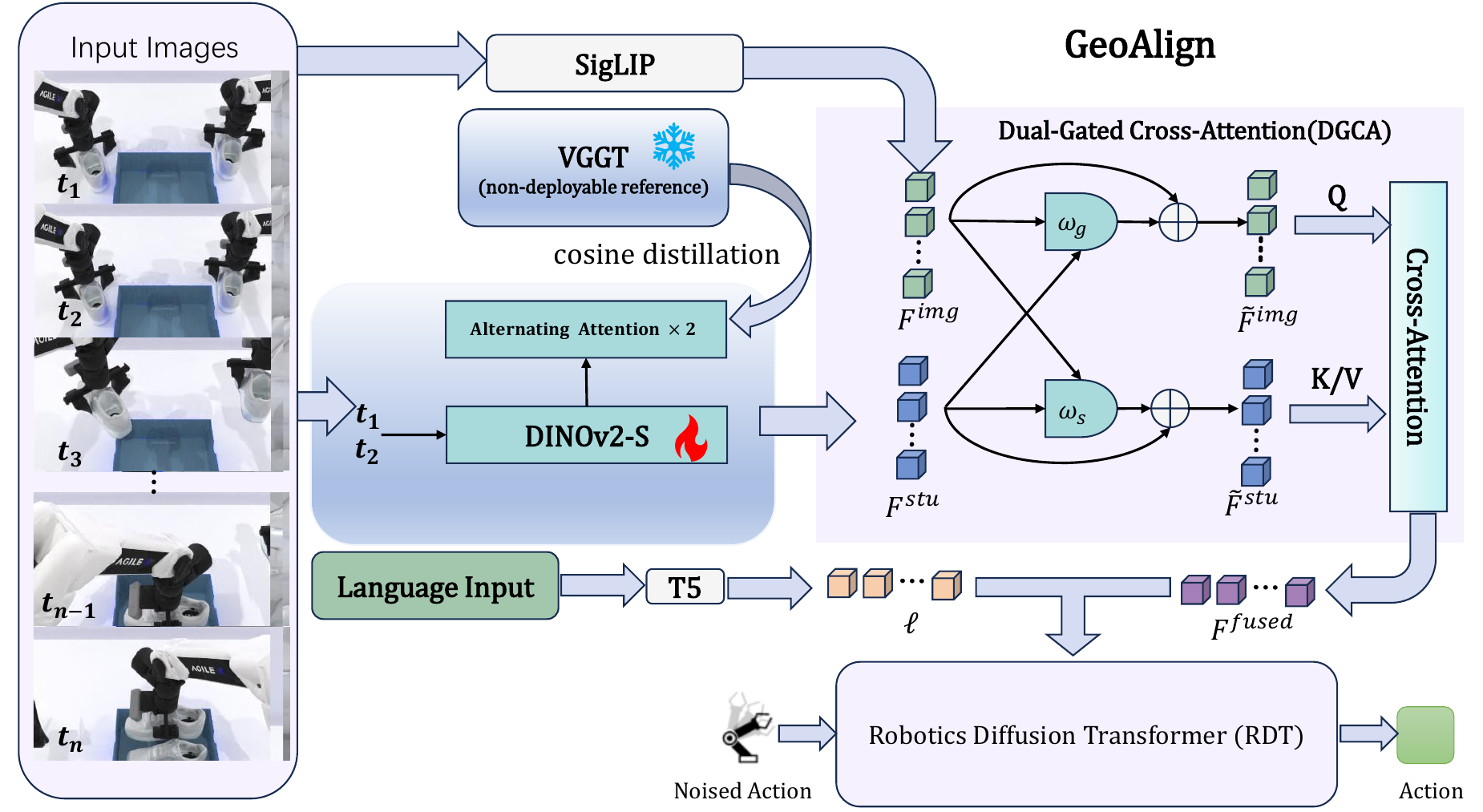}
\caption{\figurecaptiontitle{GeoAlign framework overview} Given RGB observations,
GeoAlign distills the VGGT teacher representations into efficient, patch-aligned
geometric tokens. 
DGCA serves as the fusion module within GeoAlign, adaptively
integrating these tokens with the SigLIP image features that condition RDT.
% 这句话不要了 RDT's language, proprioceptive, and diffusion action-generation pathways are retained; the proprioceptive inputs are omitted for clarity.
}
\label{fig:overview}
\end{figure}

As illustrated in Fig.~\ref{fig:overview}, we propose GeoAlign, a lightweight
geometry-aware conditioning framework based on Robotics Diffusion Transformer (RDT).
GeoAlign employs an efficient geometry tokenizer distilled from VGGT to
extract patch-aligned geometric tokens.
DGCA adaptively integrates these tokens with the visual features supplied to
RDT for geometry-aware action generation.
%
%The rest of this section is organized as follows: Sec.~\ref{sec:preliminary} reviews the preliminaries of RDT and VGGT. 
%
%Sec.~\ref{sec:geoalign} and~\ref{sec:dgca} then introduce GeoAlign and DGCA, respectively.
%
%Finally, Sec.~\ref{sec:training} describes the training procedure and objective.

\subsection{Preliminary}
\label{sec:preliminary}

\runinheading{RDT action generation.}
We build on the RDT~\cite{liu2025rdt} for language-conditioned bimanual manipulation.
At control step $t$, RDT receives a language instruction $\ell$ and an
observation $o_t=(\mathcal{X}_t,z_t,c)$, where $\mathcal{X}_t$ is the RGB
observation history, $z_t$ is the robot proprioceptive state, and $c$ is the
control frequency. It models the conditional distribution of the continuous
action chunk $a_{t:t+T_a}=(a_t,\ldots,a_{t+T_a-1})$, where $a_t$ is the robot
action at step $t$ and $T_a$ is the action horizon, with a diffusion Transformer
rather than directly regressing a deterministic action.
The denoiser $f_\theta$, parameterized by $\theta$, is trained to recover the
clean action chunk from its noised version, conditioned on $\ell$, $o_t$, and
diffusion step $k$:
\begin{equation}
\begin{aligned}
\widetilde{a}_{t:t+T_a} &= \sqrt{\bar{\alpha}_k}a_{t:t+T_a}
 + \sqrt{1-\bar{\alpha}_k}\,\epsilon,\\
L_{\mathrm{RDT}} &= \mathrm{MSE}\!\left(
a_{t:t+T_a}, f_\theta(\ell,o_t,\widetilde{a}_{t:t+T_a},k)
\right).
\end{aligned}
\label{eq:rdt}
\end{equation}
Here, $\epsilon\sim\mathcal{N}(0,I)$ is Gaussian noise with identity covariance,
$\bar{\alpha}_k$ is the cumulative noise-schedule coefficient at step $k$, and
$\widetilde{a}_{t:t+T_a}$ is the resulting noised action chunk.
$L_{\mathrm{RDT}}$ is the mean-squared-error action-generation loss used to
train the RDT denoiser to recover the clean continuous action chunk.
At inference, RDT starts from Gaussian action noise and iteratively denoises it into an executable action chunk.
We use RDT as our base model and retain this action-generation process unchanged.

\runinheading{Visual geometry encoding of VGGT.}
VGGT is a feed-forward visual foundation
model that maps a sequence of $N$ RGB images to camera parameters, depth maps,
point maps, and dense tracking features:
\begin{equation}
f_{\mathrm{VGGT}}\!\left((I_i)_{i=1}^{N}\right)
=\left(g_i,D_i,P_i,T_i\right)_{i=1}^{N}.
\label{eq:vggt}
\end{equation}
Here, $I_i\in\mathbb{R}^{H\times W\times3}$ is the $i$-th RGB image, with
$H$ and $W$ denoting its spatial resolution and $N$ is the number of input images. 
The outputs $g_i$, $D_i\in\mathbb{R}^{H\times W}$,
$P_i\in\mathbb{R}^{H\times W\times3}$, and $T_i$ denote the corresponding
camera parameters, depth map, 3D point map, and dense tracking representation,
respectively.
Each image is patchified into tokens, which are processed by alternating frame-wise and global self-attention to encode both local appearance and scene-level geometry.
We use the final image-token features as geometry-aware teacher representations for robot observations.
%
%Because running the full model at every control step is too costly for real-time control, GeoAlign distills these representations into a lightweight geometry tokenizer.

\subsection{Distilling VGGT into a Geometry Tokenizer}
\label{sec:geoalign}

\runinheading{Teacher and student feature computation.}
The VGGT aggregator produces patch-level features that implicitly encode camera geometry and scene structure.
However, applying the aggregator to every frame is computationally prohibitive for real-time control.
We therefore use the output image-token features from its final block as the distillation targets.
This feature-level supervision enables the student to learn geometry-aware representations without reproducing VGGT’s camera, depth, and point-map prediction heads. The student is thus optimized as a compact geometry tokenizer whose outputs provide geometric conditioning for the downstream policy.
Given a synchronized multi-view sequence $\mathcal{I}_t=(I_{t,i})_{i=1}^{N}$, the teacher jointly processes all views and produces per-view, per-patch features $F_{t,i}^{\mathrm{tea}}=[h_{\mathrm{VGGT}}(\mathcal{I}_t)]_i \in\mathbb{R}^{M\times d_t}$, where $M$ denotes the number of patch tokens per view, $d_t$ is the teacher feature dimension, and $h_{\mathrm{VGGT}}$ denotes the teacher's final-block image-token mapping.
To efficiently approximate these geometry-aware representations, the student comprises a trainable DINOv2-S~\cite{oquab2023dinov2} encoder, a learnable feature projection, and two alternating-attention blocks.
Specifically, the DINOv2-S encoder first converts each view into 384-dimensional patch tokens, after which a learnable linear projection maps them to the student geometry dimension $d_g$.
The alternating-attention blocks then apply frame-wise self-attention within each view and global self-attention across all views, allowing the student to capture both within-view spatial structure and cross-view geometric context, producing $F_{t,i}^{\mathrm{stu}}=[f_{\mathrm{stu}}(\mathcal{I}_t)]_i \in\mathbb{R}^{M\times d_g}$ for each view $i$.
We set $d_g=d_t$ so that the projected student tokens and teacher targets share the same feature dimension.
Omitting the frame index $t$ for brevity, the student learns to match the teacher features using the following cosine-similarity objective:
\begin{equation}
L_{\mathrm{geo}}=\frac{1}{N}\sum_{i=1}^{N}
\left(1-\cos(F_i^{\mathrm{stu}},F_i^{\mathrm{tea}})\right).
\label{eq:geo}
\end{equation}

\runinheading{Patch alignment.}
The geometric features produced by the student and the visual features extracted by SigLIP~\cite{zhai2023sigmoid} are defined on patch grids of different spatial resolutions, resulting in mismatched token counts.
Therefore, we resize each view's two-dimensional student feature grid to match the spatial resolution of its corresponding SigLIP patch grid.
The resized features are then flattened into geometric tokens, yielding the same number of tokens as the corresponding SigLIP representation and establishing view-wise spatial correspondence between the two modalities.

\subsection{Dual-Gated Cross-Attention Fusion}
\label{sec:dgca}

\begin{comment}
\runinheading{Bidirectional gating.}
Let $F^{\mathrm{img}}\in\mathbb{R}^{M\times d_s}$ denote the SigLIP image tokens and $F^{stu}\in\mathbb{R}^{M\times d_g}$ denote the geometric tokens produced by the student.
%
Because both token sequences are defined on the same spatial grid, their tokens are aligned patch by patch.
%
We construct two input-dependent gates that enable each modality to modulate the other:

\begin{equation}
w_g=\sigma(W_g G),\qquad w_i=\sigma(W_i F^{\mathrm{img}})
\label{eq:gate}
\end{equation}
where $W_g\in\mathbb{R}^{d_s\times d_g}$ and $W_i\in\mathbb{R}^{d_g\times d_s}$: the geometry-conditioned gate $w_g$ reweights the image-token stream, and the image-conditioned gate $w_i$ reweights the geometric stream.
%
The two streams modulate each other through residual connections, so that the original stream is not disrupted when either direction's modulation is zero:
\begin{equation}
\widetilde{F}^{\mathrm{img}}=F^{\mathrm{img}}\odot w_g+F^{\mathrm{img}},
\qquad \widetilde{F}^{stu}=F^{stu}\odot w_i+F^{stu}
\label{eq:mod}
\end{equation}
\end{comment}

\runinheading{Bidirectional gating.}
Let $F^{\mathrm{img}}\in\mathbb{R}^{M\times d_s}$ denote the SigLIP image tokens and $F^{\mathrm{stu}}\in\mathbb{R}^{M\times d_g}$ the geometric tokens produced by the student.
Because the two token sequences share the same spatial grid, they are aligned patch by patch.
We construct two input-dependent gates that allow the geometry and image modalities to modulate each other:
\begin{equation}
w_g=\sigma\left(F^{\mathrm{stu}}W_g\right)
\in\mathbb{R}^{M\times d_s},
\quad
w_s=\sigma\left(F^{\mathrm{img}}W_s\right)
\in\mathbb{R}^{M\times d_g},
\label{eq:gate}
\end{equation}
where $W_g\in\mathbb{R}^{d_g\times d_s}$ and
$W_s\in\mathbb{R}^{d_s\times d_g}$ are learnable projections.
Specifically, the geometry-conditioned gate $w_g$ modulates the image-token stream, whereas the image-conditioned gate $w_s$ modulates the geometric-token stream.
We apply the two gates through residual modulation:
\begin{equation}
\widetilde{F}^{\mathrm{img}}=F^{\mathrm{img}}+F^{\mathrm{img}}\odot w_g,
\quad
\widetilde{F}^{\mathrm{stu}}=F^{\mathrm{stu}}+F^{\mathrm{stu}}\odot w_s,
\label{eq:mod}
\end{equation}
where $\odot$ denotes element-wise multiplication.

\begin{table}[!t]
\centering
\tablebodyfont
\caption{\tablecaptiontitle{Simulation task success rates} Results over 100
domain-randomized episodes per task. Dark-blue parenthesized values show the
change relative to RDT for the GeoAlign variants.
{\setlength{\fboxsep}{0.5pt}\colorbox{BestCell}{Green}} and
{\setlength{\fboxsep}{0.5pt}\colorbox{SecondCell}{orange}} denote the best and second-best deployable
models, respectively; the non-deployable Teacher is excluded from ranking.}
\label{tab:main}
\setlength{\tabcolsep}{3.5pt}
\begin{tabular*}{\textwidth}{@{\extracolsep{\fill}}l|cccccc|c}
\toprule
\textbf{Task} & \textbf{$\pi_0$} & \textbf{DP3} & \textbf{Spatial Forcing} & \textbf{RDT} & \textbf{\oursbase} & \textbf{\oursfull} & \textbf{Teacher} \\
\midrule
Beat Block Hammer     & 0.25 & 0.08 & 0.06 & 0.21 & \second{\gain{0.32}{+0.11}} & \best{\gain{0.33}{+0.12}} & 0.36 \\
Dump Bin BigBin       & 0.33 & \second{0.53} & \best{0.56} & 0.27 & \gain{0.27}{+0.00} & \gain{0.41}{+0.14} & 0.47 \\
Place Container Plate & \best{0.38} & 0.01 & 0.15 & 0.10 & \gain{0.12}{+0.02} & \second{\gain{0.19}{+0.09}} & 0.20 \\
Place Dual Shoes      & 0.02 & 0.02 & \second{0.04} & 0.00 & \gain{0.03}{+0.03} & \best{\gain{0.06}{+0.06}} & 0.08 \\
\midrule
Average               & \second{0.245} & 0.160 & 0.203 & 0.145 & \gain{0.185}{+0.040} & \best{\gain{0.248}{+0.103}} & 0.278 \\
\bottomrule
\end{tabular*}
\end{table}

\runinheading{Cross-stream attention.} For clarity, we omit the view index below;
the same operation is applied independently to each aligned view. A
cross-attention layer lets each SigLIP image token retrieve geometric evidence
across all spatial positions within that view:
\begin{equation}
\begin{gathered}
Q=\widetilde{F}^{\mathrm{img}}W_Q,\quad
K=\widetilde{F}^{\mathrm{stu}}W_K,\quad
V=\widetilde{F}^{\mathrm{stu}}W_V,\\
F^{\mathrm{attn}}=\mathrm{softmax}\!\left(
\frac{QK^{\top}}{\sqrt{d_a}}\right)V
\end{gathered}
\label{eq:attn}
\end{equation}
where $W_Q\in\mathbb{R}^{d_s\times d_a}$,
$W_K,W_V\in\mathbb{R}^{d_g\times d_a}$, and $d_a$ is the attention
dimension. An output projection $W_O\in\mathbb{R}^{d_a\times d_s}$ followed by
a residual connection and layer normalization yields the fused tokens:
\begin{equation}
F^{\mathrm{fused}}=\mathrm{LN}\!\left(
F^{\mathrm{attn}}W_O+\widetilde{F}^{\mathrm{img}}\right).
\label{eq:fused}
\end{equation}

\subsection{Training}
\label{sec:training}
Our training procedure consists of two stages.
Stage I learns a geometry tokenizer shared across tasks using all demonstrations, whereas Stage II freezes the tokenizer and fine-tunes the downstream action policy.

% Declare the wide main-results table early enough for stfloats to place it at
% the bottom of page 4, before the later page-wide rollout figure.  It remains
% an independent table float; its discussion is retained in Sec. IV-B.

\runinheading{Stage I (geometry distillation).}
For each sampled frame, synchronized RGB observations from the main camera and one or two wrist cameras are jointly processed as a multi-view sequence.
The VGGT teacher is frozen, while the entire geometry tokenizer, including the DINOv2-S encoder, feature projection, and two alternating-attention blocks, is optimized using the distillation objective in Eq.~\eqref{eq:geo}.

\runinheading{Stage II (policy fine-tuning). }
After distillation, the VGGT teacher is discarded and the geometry student is frozen.
We optimize only \oursfusion, the image-token projection, and the RDT denoising network end-to-end using the action-chunk denoising objective in Eq.~\eqref{eq:rdt}.
The language, proprioceptive, control-frequency, and diffusion-timestep inputs retain the original RDT formulation, whereas $F^{\mathrm{fused}}$ replaces the original SigLIP image tokens.
At inference, the cached text tokens and online-fused visual tokens jointly condition RDT, which iteratively denoises each continuous action chunk and replans upon receiving the next observation.

\section{EXPERIMENTS}

\subsection{Implementation Details}
\label{sec:implementation}
We implement GeoAlign with a frozen VGGT teacher and a geometry student comprising a trainable DINOv2-S encoder, a learnable linear projection from 384 to 2048 dimensions, and two attention blocks.
Both networks process images at a resolution of $518\times518$, and their output features are adaptively pooled to a $27\times27$ patch grid to match the spatial layout of the SigLIP image tokens.
Each aligned feature sequence therefore contains $M=27^2=729$ tokens.
The teacher and student geometry dimensions are $d_t=d_g=2048$, while the SigLIP image-token dimension is $d_s=1152$.
The student contains 74M parameters, all of which are optimized during Stage I.
During policy training, language embeddings are precomputed and cached offline~\cite{raffel2020exploring}.
Each policy step processes observations from three cameras over two history frames.
Specifically, these two frames consist of the previous frame and the current frame, which are stacked as input to the policy network.
During Stage II policy training and inference, the distilled geometry tokenizer is frozen and evaluated under $\mathrm{no\_grad}$.
Language, visual, and low-dimensional adapters project their inputs into RDT's 2048-dimensional token space.
The RDT denoiser predicts 64-step action chunks at a 25 Hz control frequency on an RTX PRO 6000 GPU.

\subsection{Simulation Setup}
We evaluate on four \dataset~\cite{mu2025robotwin, chen2025robotwin} tasks on the AgileX dual-arm platform in simulation, built on SAPIEN~\cite{xiang2020sapien}, following common simulation benchmarks: \textit{Beat Block Hammer}, \textit{Dump Bin BigBin}, \textit{Place Container Plate}, and \textit{Place Dual Shoes}.
All policies are evaluated over 100 seeded episodes in the domain-randomized configuration: backgrounds and lighting are randomized, table height varies by $\pm$3 cm, tables are cluttered with distractors, and clean-background scenes occur with probability 0.02.
Observations comprise RGB images from head and wrist D435 cameras, joint states, and end-effector poses.
We report task success rates.
We compare the following baselines: \textit{1) RDT}, the 2D RDT baseline with SigLIP only; \textit{2) \teacher}, which uses full VGGT features with concatenation fusion as a non-deployable accuracy reference; \textit{3) \oursbase}, distilled features with only concatenation fusion; and \textit{4) \oursfull}, distilled features with our DGCA fusion; \textit{5) $\pi_0$}~\cite{black2024pi_0} fine-tuned with LoRA.

\begin{table}[!t]
\centering
\tablebodyfont
\caption{\tablecaptiontitle{Student--teacher feature alignment} Cosine distance
($1-\cos$; lower is better) between student and teacher features, averaged over
50 frames per task.}
\label{tab:fidelity}
\setlength{\tabcolsep}{3pt}
\begin{tabular*}{\columnwidth}{@{\extracolsep{\fill}}lccc@{}}
\toprule
\textbf{Task} & \textbf{Main cam} & \textbf{Wrist cams} & \textbf{Multi-view} \\
\midrule
Beat Block Hammer      & 0.054 & 0.118 & 0.097 \\
Dump Bin BigBin        & 0.037 & 0.121 & 0.093 \\
Place Container Plate  & 0.048 & 0.164 & 0.125 \\
Place Dual Shoes       & 0.035 & 0.112 & 0.086 \\
\midrule
Average                & 0.044 & 0.129 & 0.100 \\
\bottomrule
\end{tabular*}
\end{table}

\subsection{Main Results}
\label{sec:main}
\begin{figure}[!htbp]
\centering
\includegraphics[width=1\textwidth]{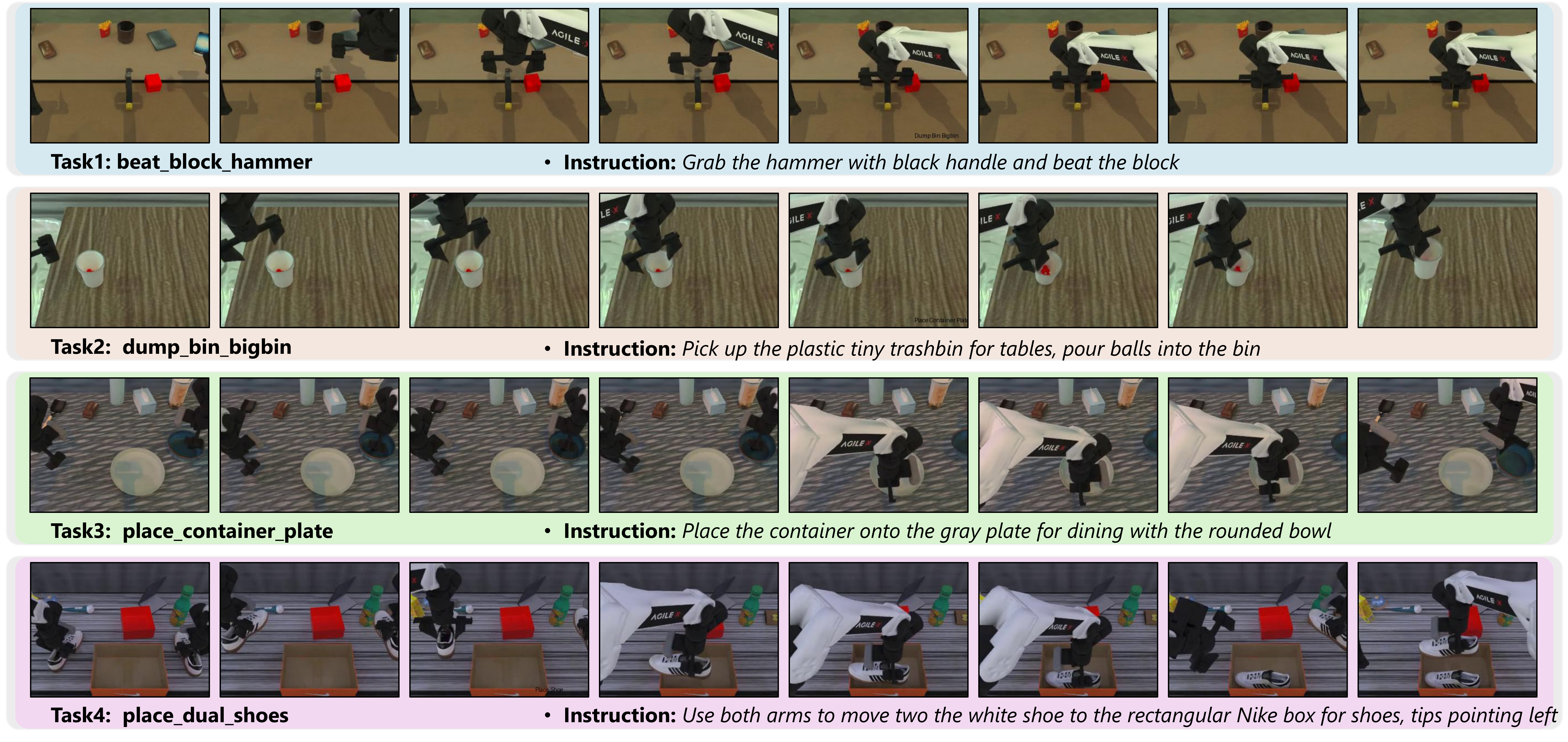}
\caption{\figurecaptiontitle{Simulation rollout sequences} Each row gives the task name and an unseen instruction, followed by eight keyframes spanning a successful rollout.}
\label{fig:rollout}
\end{figure}

Table~\ref{tab:main} summarizes the task-wise success rates across the four simulation tasks.
The non-deployable \teacher achieves an average success rate of 0.278, compared
with 0.145 for the 2D-only RDT baseline. Because both use the same RDT action
generation, this difference shows that VGGT geometry provides useful spatial
conditioning for manipulation.

\oursbase reaches 0.185, improving over RDT by 0.040. This gain confirms that
the lightweight tokenizer preserves geometric information useful for control.
With the same concatenation fusion, replacing full VGGT with the student tokenizer reduces the average success rate from 0.278 to 0.185, a drop of 0.093.

Adding DGCA raises the average from 0.185 to 0.248 and improves all four tasks, recovering 67.7\% of the gap between \oursbase and the teacher while retaining 89.2\% of the teacher's success rate.
This improvement comes from more effective semantic--geometric fusion.

Among deployable methods, \oursfull obtains the highest average success rate, at 0.248 versus 0.245 for $\pi_0$ and 0.203 for Spatial Forcing~\cite{li2026spatial}, and outperforms each baseline on three of four tasks.
This result shows that efficient geometry distillation and adaptive fusion translate into competitive task performance.
Fig.~\ref{fig:rollout} shows representative successful rollouts under unseen
instructions.

\subsection{Distillation Performance}
\label{sec:distillation_performance}

% Declare Table III at the start of Sec. C so it enters the independent table
% queue before the later qualitative figure enters the figure queue.
\begin{table}[!t]
\centering
\tablebodyfont
\caption{\tablecaptiontitle{Latency breakdown}  BF16 inference latency (ms) on RTX PRO 6000,
averaged over 20 runs. Obs encode $=$ SigLIP extraction $+$ geometry
extraction $+$ fusion.}
\label{tab:latency}
\setlength{\tabcolsep}{2.5pt}
\begin{tabular*}{\columnwidth}{@{\extracolsep{\fill}}lcccc@{}}
\toprule
\textbf{Component} & \textbf{RDT} & \textbf{GeoAlign} & \textbf{GeoAlign} & \textbf{Teacher} \\
                   &              & \textbf{w/o DGCA} &                   &                  \\
\midrule
SigLIP extract & 52.9 & 52.9 & 52.9 & 52.9 \\
Geo extract    & -- & \best{15.4} & \best{15.4} & 137.0 \\
Fusion         & -- & \best{0.5}  & 3.8  & \best{0.5} \\
\midrule
Obs encode     & \best{52.9} & \second{68.8} & 72.1 & 190.4 \\
Per-chunk inference  & 277.8 & 353.8 & 364.7 & 1285.8 \\
\bottomrule
\end{tabular*}
\end{table}

\begin{wrapfigure}{R}{0.47\textwidth}
\centering
\includegraphics[width=\linewidth]{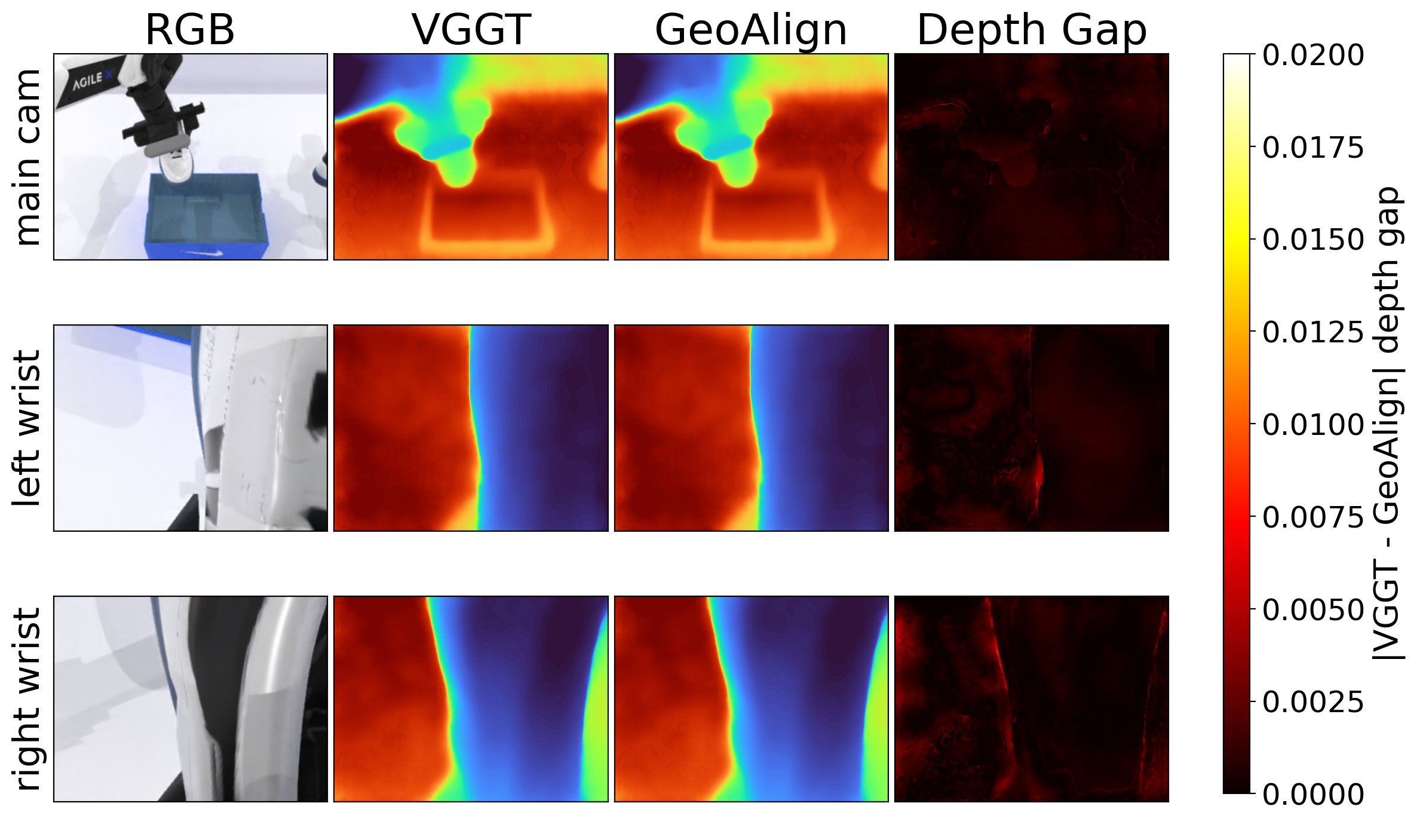}
\caption{\figurecaptiontitle{Qualitative distillation performance} Depth maps decoded from teacher (VGGT) and student (\ours) features using the same frozen teacher depth head.
The student preserves the overall scene geometry in all three views.}
\label{fig:depth}
\end{wrapfigure}
We next examine whether the lightweight tokenizer faithfully preserves the VGGT teacher's geometric representations across the main, left-wrist, and right-wrist views.
As shown in Fig.~\ref{fig:depth}, we decode depth maps from the teacher and student features using the same frozen teacher depth head.
The resulting depth maps are highly consistent across all three views, which demonstrates that the geometric knowledge of the VGGT teacher is effectively transferred to the lightweight tokenizer.
For a quantitative evaluation, we compute the cosine distance between the student and teacher last-block features at every patch, and then average the distances over patches, frames, and tasks.
As reported in Table~\ref{tab:fidelity}, the average student--teacher cosine distances are 0.044 for the main camera, 0.129 for the wrist cameras, and only 0.100 across all views.
Together, these qualitative and quantitative results show that the student maintains feature alignment across all three camera views.

% Figure 4 remains an independent top float declared after Table III.

\par
\subsection{Efficiency}
\label{sec:efficiency}

Table~\ref{tab:latency} reports the latency breakdown for observation encoding
and action decoding.
%
%Because all variants share the same SigLIP extraction pathway, the principal difference lies in geometry extraction and fusion.
%
The student requires only 15.4 ms for geometry extraction, compared with 137.0 ms for the VGGT teacher, an 8.9$\times$ speedup.
\oursfusion requires 3.8 ms, adding only 3.3 ms over concatenation.
Including the shared SigLIP extraction, the complete observation-encoding latency is 72.1 ms for \oursfull and 190.4 ms for the teacher, corresponding to a 2.6$\times$ speedup.
The tokenizer therefore removes most of the per-frame geometry encoding cost, while adaptive fusion introduces only modest overhead relative to the savings from distillation.

\begin{table}[!t]
\centering
\tablebodyfont
\caption{\tablecaptiontitle{Ablation of DGCA design} CA retains only cross-attention, while Fixed-Gate uses input-independent gate parameters. All variants are independently trained and evaluated over 100 episodes per task.}
\label{tab:ca_retrain}
\setlength{\tabcolsep}{1.5pt}
\begin{tabular*}{\columnwidth}{@{\extracolsep{\fill}}lcccc@{}}
\toprule
\textbf{Task} & \textbf{GeoAlign} & \textbf{CA} & \textbf{Fixed-Gate} & \textbf{GeoAlign} \\
              & \textbf{w/o DGCA} &             &                     &                  \\
\midrule
Beat Block Hammer      & \second{0.32} & 0.08 & 0.30 & \best{0.33} \\
Dump Bin BigBin        & 0.27 & 0.20 & \second{0.36} & \best{0.41} \\
Place Container Plate  & \second{0.12} & \best{0.19} & 0.10 & \best{0.19} \\
Place Dual Shoes       & \second{0.03} & 0.01 & \second{0.03} & \best{0.06} \\
\midrule
Average                & 0.185 & 0.120 & \second{0.198} & \best{0.248} \\
\bottomrule
\end{tabular*}
\vspace{0.6\baselineskip}
\end{table}

\subsection{Ablation: Gate Design}
\label{sec:ablation}
Table~\ref{tab:ca_retrain} presents a controlled ablation of three operations in DGCA: cross-attention, gated modulation, and input-conditioned gating.
All variants are independently trained from scratch using the same data and objective.
\oursbase concatenates the semantic and geometric features; CA replaces concatenation with cross-attention but removes both gates; Fixed-Gate uses input-independent gate parameters; and the complete \oursfull model predicts gate outputs from each observation.

CA reaches an average success rate of 0.120, which is 0.065 below the 0.185 obtained by \oursbase.
In particular, CA reduces the success rate from 0.32 to 0.08 on \textit{Beat Block Hammer} and from 0.27 to 0.20 on \textit{Dump Bin BigBin}, despite improving \textit{Place Container Plate} from 0.12 to 0.19.
This uneven change indicates that cross-attention alone does not consistently exploit the distilled geometric tokens; therefore, the gain of DGCA cannot be attributed merely to adding an attention layer, but depends on gated regulation of semantic--geometric interaction.

Introducing input-independent modulation raises the average success rate from 0.120 for CA to 0.198 for Fixed-Gate, including improvements from 0.08 to 0.30 on \textit{Beat Block Hammer} and from 0.20 to 0.36 on \textit{Dump Bin BigBin}.
This 0.078 gain shows that gated modulation is beneficial even without input conditioning.
The complete \oursfull model further improves all four tasks over Fixed-Gate, raising the respective success rates from 0.30, 0.36, 0.10, and 0.03 to 0.33, 0.41, 0.19, and 0.06, and increasing the average from 0.198 to 0.248.
The additional 0.050 gain supports the input-adaptive design of DGCA, which adjusts the relative contribution of semantic and geometric features according to the current observation rather than applying the same modulation to every input.

\subsection{Real-World Experiments}
\label{sec:realworld}

\par\addvspace{8pt}
\noindent\begin{minipage}[t]{0.49\textwidth}
\vspace{0pt}
We evaluate \ours on a Hiwonder SO-101 dual-arm platform, following the low-cost dual-arm teleoperation paradigm of Mobile ALOHA~\cite{fu2024mobile}.
Fig.~\ref{fig:realworld_setup} illustrates the real-world setup, which consists of one fixed camera as the main view and two robot arms with wrist-mounted cameras.

\runinheading{Task Description.}  \textit{Pick Place Block} requires the policy to localize a randomly placed block of the instructed color and place it on the target plate.
Low foreground--background contrast and illumination-induced appearance variations weaken visual cues, making robust geometric reasoning essential for accurate grasping and placement.

% \textit{Pick Place Block} evaluates both instruction understanding and 3D spatial perception.
% %
% Given a language instruction specifying a block color, the policy must identify the corresponding block and estimate the 3D location, including depth, of an appropriate grasp point before placing the block on the target plate.
% %
% The semantic features support color-conditioned target identification, while the distilled geometry tokens provide complementary spatial cues for 3D localization and grasping.
% %
% DGCA adaptively integrates these two sources of information for each input.

% \textit{Stack Bowls} places greater emphasis on precise 3D alignment.
% %
% The policy must localize the red and white bowls, grasp the red bowl, align it above the white bowl, and execute a controlled vertical descent and release.
% %
% Success therefore depends on both the semantic roles of the two bowls and their
% relative depth, height, and center alignment.
% %
% %Across both tasks, observations transition from global localization in the fixed-camera view to close-range manipulation in the wrist views.
% %
% GeoAlign supplies geometry in each view, and DGCA independently calibrates the semantic and geometric streams for each input.

\textit{Stack Bowls} places greater emphasis on precise 3D alignment.
The policy must distinguish the red and white bowls and align the red bowl with the center of the white bowl before release.
Success therefore depends on estimating relative depth, height, and center alignment.
Together, the two tasks evaluate semantic target selection and geometry-aware localization across the fixed and wrist-camera observations.
\end{minipage}\hfill%
\begin{minipage}[t]{0.47\textwidth}
\vspace{0pt}

\centering
% The source slide has generous outer whitespace; crop it so the visual itself
% fills the column while preserving its aspect ratio and native PDF quality.
\includegraphics[width=1\linewidth]{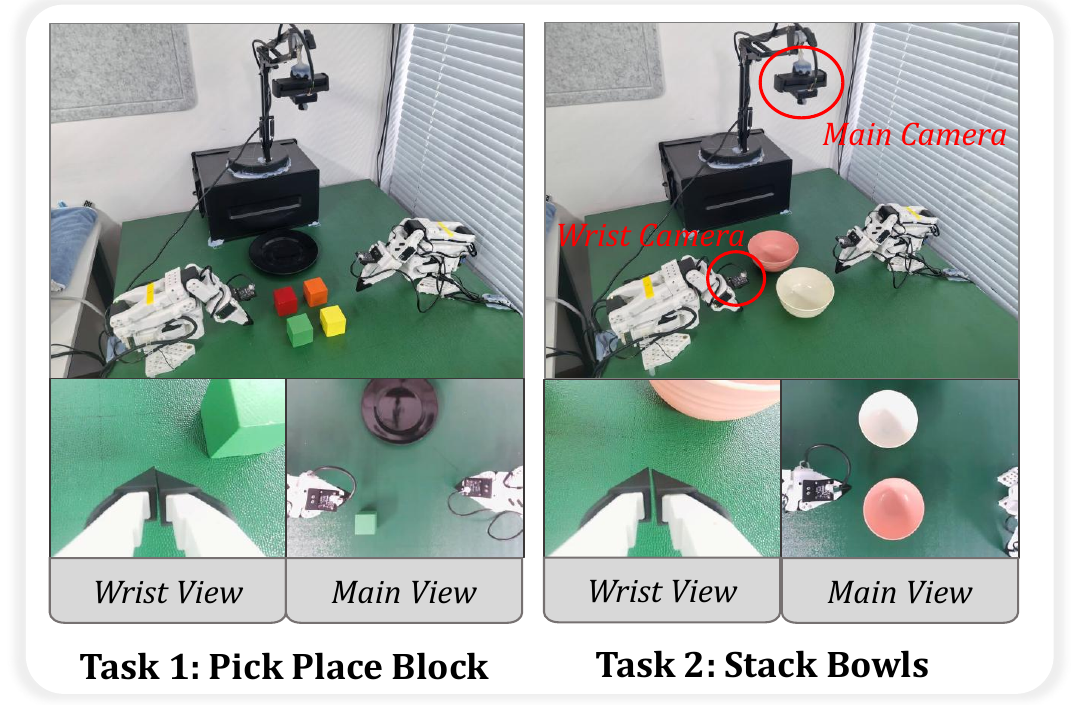}
\captionof{figure}{\figurecaptiontitle{Real-world platform and tasks} The Hiwonder SO-101
dual-arm robot. The figure identifies the fixed main and wrist cameras, the task
object layout for tasks \textit{Pick Place Block} and \textit{Stack Bowls}.}
\label{fig:realworld_setup}

\end{minipage}\par\addvspace{8pt}
\par

\begin{figure}[!htbp]

\noindent\begin{minipage}[t]{0.49\textwidth}
\vspace{0pt}
\runinheading{Data collection.} For \textit{Pick Place Block}, we collect 40
demonstrations for each of four block colors, for 160 demonstrations in total.
The block position varies substantially across demonstrations, while the plate
position varies only slightly.
For \textit{Stack Bowls}, the robot stacks a red bowl inside a white bowl, with
both bowl positions randomized across 120 demonstrations.

\runinheading{Evaluation.} We use the same training setting as in the simulation benchmark.
For \textit{Pick Place Block}, one of the four colors is selected, and the block is placed at a random position in each trial.
For \textit{Stack Bowls}, both bowl positions are re-randomized in every trial, and each task is evaluated over 20 trials.
All four RDT variants are evaluated together with the $\pi_0$, DP3, and the teacher-based variant, which serves as an accuracy reference.
\end{minipage}\hfill%
\begin{minipage}[t]{0.47\textwidth}
\vspace{0pt}

\centering
\includegraphics[width=\linewidth]{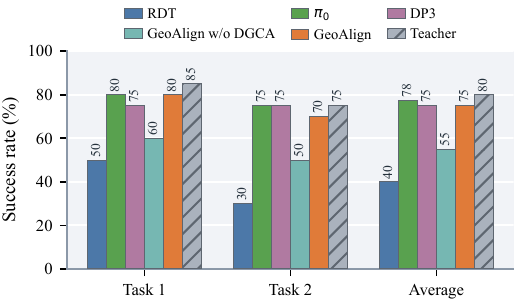}
\caption{\figurecaptiontitle{Real-world task success rates} Task 1 is Pick Place Block and Task 2 is Stack Bowls; each method is evaluated over 20 trials per task. Average is their mean, and the hatched bars denote the non-deployable teacher reference.}
\label{fig:realworld_success}

\end{minipage}
\end{figure}

\noindent Fig.~\ref{fig:realworld_success} compares the two task success rates and their
average.
%
%On \textit{Pick Place Block}, RDT, \oursbase, and \oursfull obtain 0.50, 0.60, and 0.80, respectively.
%
%On \textit{Stack Bowls}, their success rates are 0.30, 0.50, and 0.70.
%
The average therefore rises from 0.40 for RDT to 0.55 for \oursbase, showing that distilled geometry improves real-world spatial conditioning, and further to 0.75 for \oursfull, showing that adaptive fusion uses these features more effectively than concatenation.
\oursfull matches DP3, remains within 0.025 of $\pi_0$, and is only 0.05
below the non-deployable teacher average of 0.80.
The real-world results thus mirror the simulation trend: the geometry tokenizer provides efficient spatial conditioning across the three views, while DGCA improves the policy's utilization of the distilled geometric representations.

\begin{figure}[!htbp]
\centering
\includegraphics[width=1\textwidth]{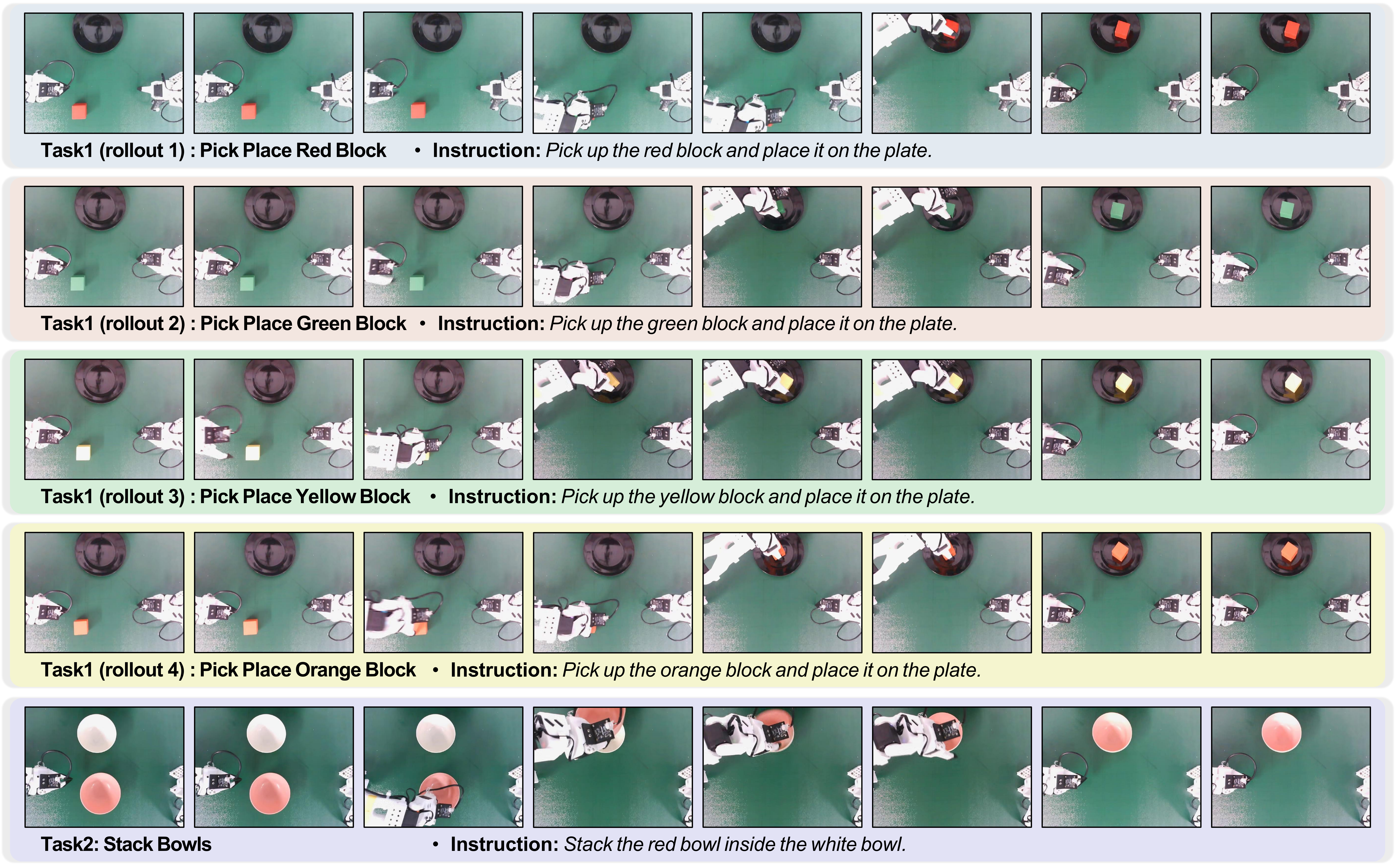}
\caption{\figurecaptiontitle{Real-world rollout sequences} Five successful rollouts, each visualized using eight temporally ordered keyframes.
The task name and language instruction are shown below each row.}
\label{fig:realworld_rollout}
\end{figure}

\runinheading{Real-world Inference Latency.} Per-chunk inference latency on the real platform is reported in Table~\ref{tab:realworld_latency}.
GeoAlign achieves a per-chunk latency of 415.1 ms, comparable to the 393.4 ms of the fastest method, $\pi_0$, which predicts an action chunk of 50 steps.
Meanwhile, it is approximately 3.3$\times$ faster than the teacher, which requires 1387.6 ms per chunk.

\begin{table}[!htbp]
% This table is embedded below Fig. 6 inside the same top-float container.
% Switching \@captype preserves table numbering and table-caption formatting.
\begingroup
\makeatletter
\def\@captype{table}
\makeatother
\centering
\tablebodyfont
\caption{\tablecaptiontitle{Real-world inference latency} The BF16 latency is averaged over
20 runs; Per-chunk inference is the latency of one 64-step action chunk, and
lower is better.}
\label{tab:realworld_latency}
\setlength{\tabcolsep}{3pt}
\begin{tabular*}{\columnwidth}{@{\extracolsep{\fill}}lccc@{}}
\toprule
\textbf{Model} & \textbf{Obs encode} & \textbf{Per-chunk inference} & \textbf{chunk length} \\
\midrule
GeoAlign w/o DGCA & 68.8  & 415.0 & 64\\
GeoAlign          & 72.1 & 415.1 & 64\\
Teacher           & 190.4 & 1387.6 & 64\\
DP3               & 131.3 & 483.4 & 6\\
$\pi_0$               &107.5     &  393.4 & 50 \\
\bottomrule
\end{tabular*}
\endgroup

\end{table}

\section{CONCLUSION}

We presented \ours{}, a lightweight geometry-aware conditioning framework that provides RDT with VGGT-derived spatial priors using only RGB observations.
By efficiently distilling geometric representations and adaptively integrating them with semantic visual features, \ours{} preserves geometric structure across main and wrist views while requiring only 37.9\% of the teacher's observation-encoding time.
Overall, \ours{} improves the average simulation success rate of RDT from 0.145 to 0.248 and achieves a 75.0\% average success rate on real-world bimanual tasks, demonstrating an effective accuracy--efficiency trade-off for geometry-aware robotic manipulation.

\bibliographystyle{unsrt}
\bibliography{reference}

\end{document}